\documentclass{article}

\usepackage[nonatbib,preprint]{neurips_2024}

\usepackage[utf8]{inputenc} 
\usepackage[T1]{fontenc}    
\usepackage{hyperref}       
\usepackage{url}            
\usepackage{booktabs}       
\usepackage{amsfonts}       
\usepackage{nicefrac}       
\usepackage{microtype}      
\usepackage[table]{xcolor}
\usepackage{enumitem}
\usepackage{graphicx}
\usepackage{adjustbox}
\usepackage{wrapfig}
\usepackage{subcaption}
\usepackage{multirow}

\usepackage[toc,page,header]{appendix}
\usepackage{minitoc}
\usepackage{tcolorbox}

\newcommand{\legodataset}{Isle-Bricks}
\newcommand{\dotsdataset}{Isle-Dots}

\newcommand{\takeaways}[1]{%
  \begin{tcolorbox}[colback=gray!5!white, boxrule=0.2mm]
    #1
  \end{tcolorbox}%
}

\title{ 
  \begin{minipage}{0.1\textwidth} 
    \includegraphics[width=\linewidth,height=1.5cm]{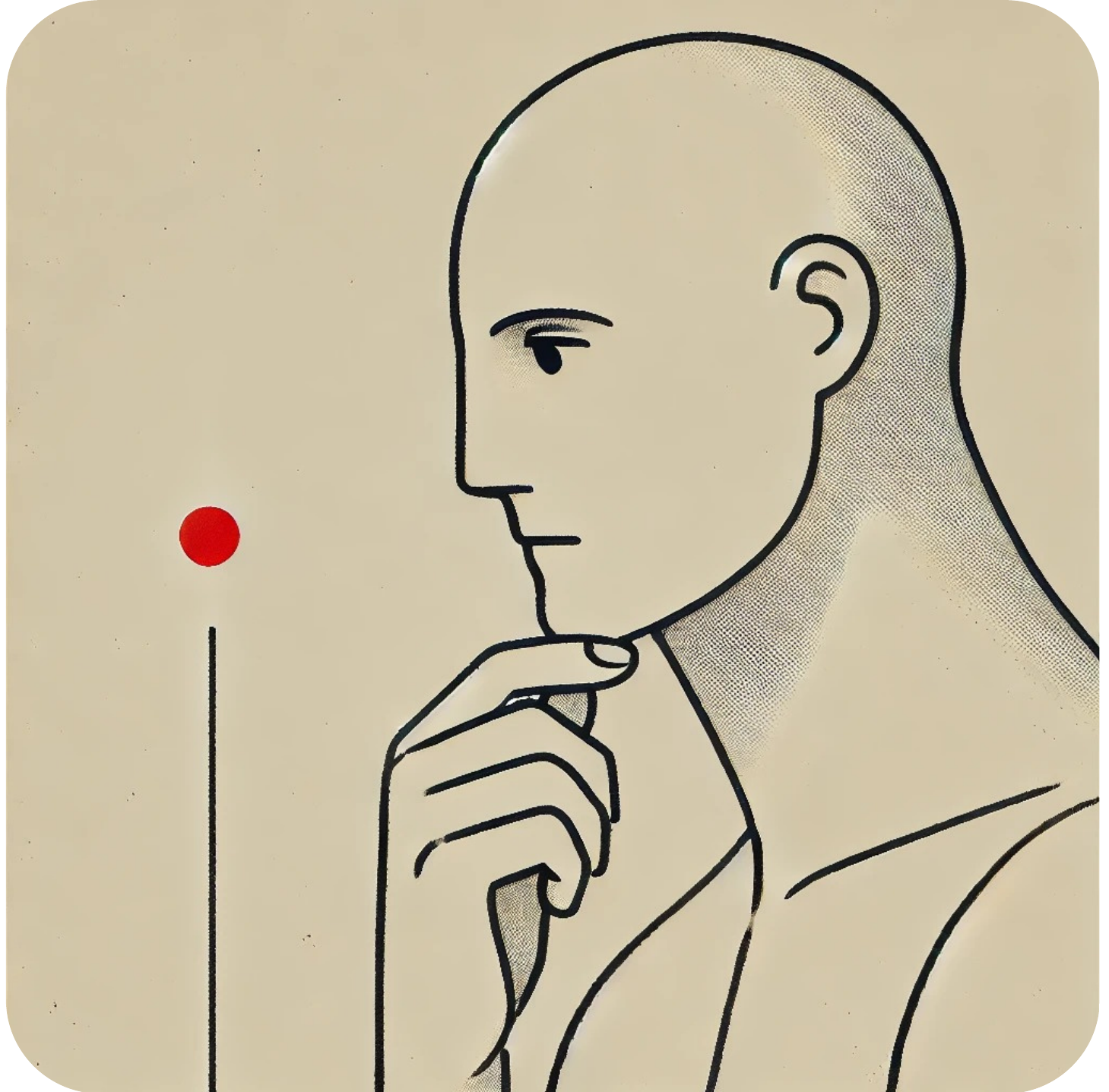} 
  \end{minipage}%
  \hspace{0.02\textwidth} 
  \begin{minipage}{0.87\textwidth} 
    \Large\bfseries Seeing Through Their Eyes: Evaluating Visual Perspective Taking in Vision Language Models
  \end{minipage}
} 

\author{%
  Gracjan Góral\thanks{University of Warsaw} \,\thanks{IDEAS NCBR} \,\thanks{IMPAN} \\
  gp.goral@uw.edu.pl \\ 
  \And
  Alicja Ziarko\footnotemark[1] \,\footnotemark[2] \,\footnotemark[3] \\
  \And
  Michał Nauman\footnotemark[1] \,\footnotemark[2] \,\thanks{Equal senior contribution}\\
  \And
  Maciej Wołczyk\footnotemark[2] \,\footnotemark[4] \\
}

\begin{document}

\definecolor{salmon}{rgb}{1, 0.55, 0.65}


\maketitle


\begin{abstract}

Visual perspective-taking (VPT), the ability to understand the viewpoint of another person, enables individuals to anticipate the actions of other people. For instance, a driver can avoid accidents by assessing what pedestrians see. Humans typically develop this skill in early childhood, but it remains unclear whether the recently emerging Vision Language Models (VLMs) possess such capability. Furthermore, as these models are increasingly deployed in the real world, understanding how they perform nuanced tasks like VPT becomes essential. In this paper, we introduce two manually curated datasets, \textit{\legodataset{}} and \textit{\dotsdataset{}} for testing VPT skills, and we use it to evaluate $12$ commonly used VLMs. Across all models, we observe a significant performance drop when perspective-taking is required. Additionally, we find performance in object detection tasks is poorly correlated with performance on VPT tasks, suggesting that the existing benchmarks might not be sufficient to understand this problem.
The code and the dataset will be available at \href{https://sites.google.com/view/perspective-taking}{this URL}.
\end{abstract}
\section{Introduction}

To navigate everyday interactions that include other people, we need to be able to imagine the world through their eyes.  
It is fundamental to avoid physical hazards (\textit{does the other driver see me on the road?}), coordinate actions effectively (\textit{is this person ready to pass me an object?}), or even respond appropriately in social settings (e.g. \textit{should I speak now?}). In psychology, visual perspective-taking (VPT) is a cognitive ability associated with viewing the world from the spatial perspective of another person~\cite{Piaget1956, Flavell1977, BaronCohen1985}, and its deficiency was linked to poor navigation~\cite{Orefice2024-oh} and social~\cite{pearson2013review} skills.
Although the recently emerging Vision Language Models (VLMs)~\cite{wang2024cogvlmvisualexpertpretrained, bubeck2023sparksartificialgeneralintelligence, aghajanyan2023scaling} exhibit impressive performance in many important tasks~\cite{geminiteam2024geminifamilyhighlycapable, openai2024gpt4technicalreport}, their VPT capabilities are largely unknown. This is concerning, as VLMs are increasingly being deployed in real-world robotic scenarios~\cite{zitkovich2023rt,kim2024openvla,michal2024robotic} that may require interaction with humans.



\begin{figure}[t!]
    \begin{center}
        \hfill
    
    \begin{minipage}[t]{0.49\linewidth}

    \begin{subfigure}{1.0\linewidth}
        \subcaption{\legodataset{}}
        \hfill
        \includegraphics[width=0.48\linewidth]{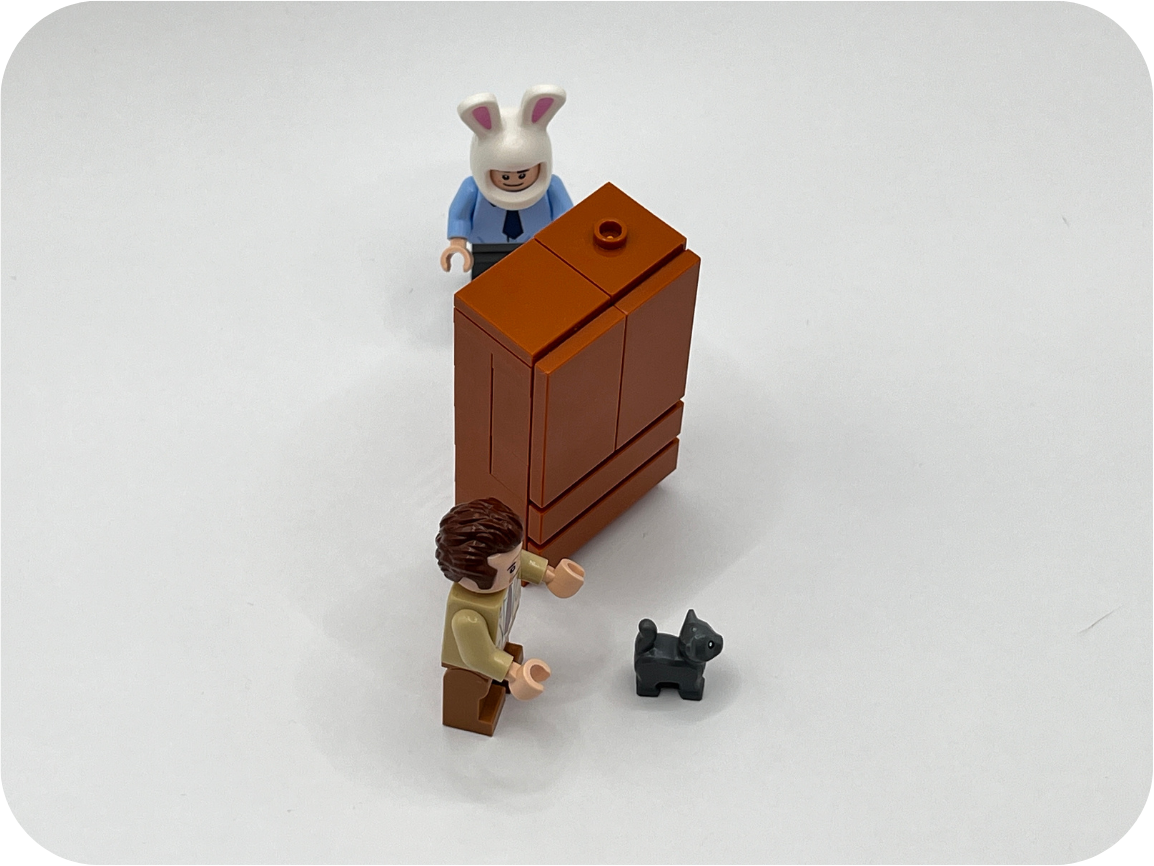}
        \hfill
        \includegraphics[width=0.48\linewidth]{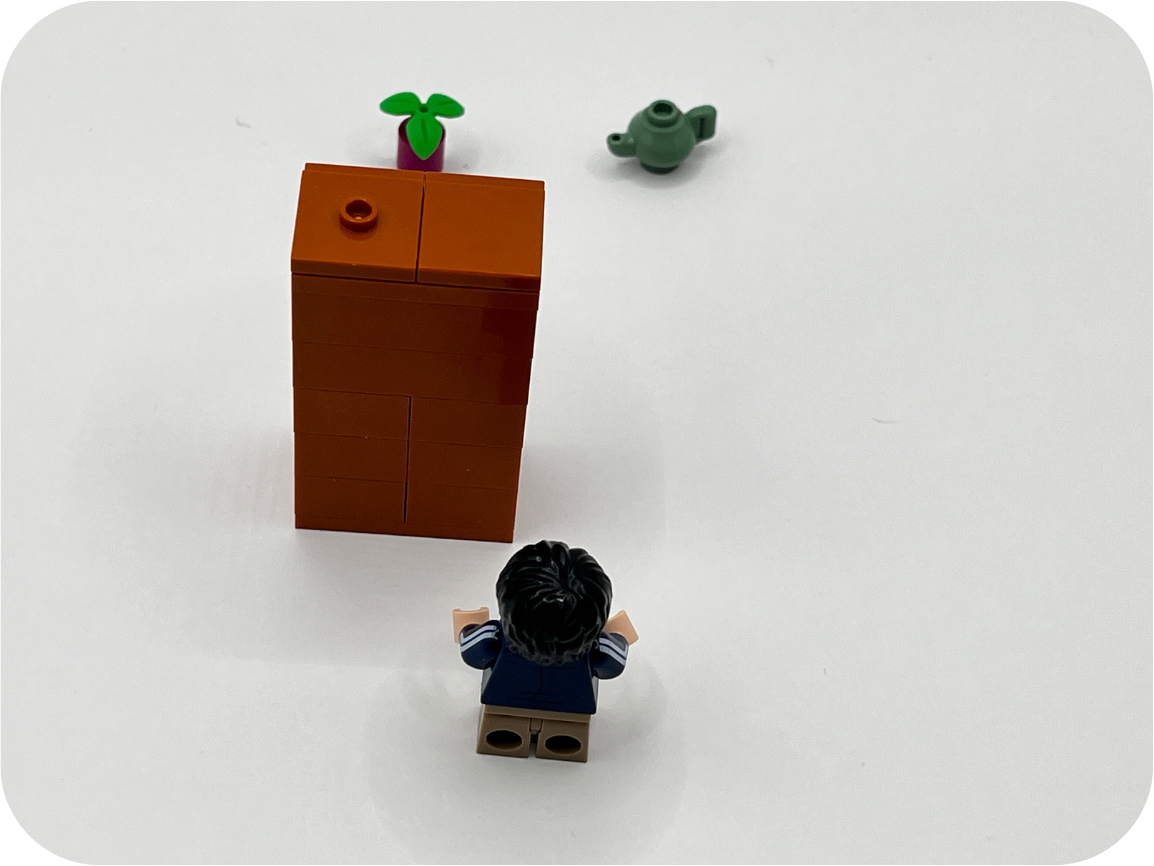}
        \hfill
    \end{subfigure}
        \centering
        \vspace{-0.2in}
        \begin{tcolorbox}[colback=gray!5!white, boxrule=0.2mm, width=0.95\linewidth]
\small \textbf{VPT:} Assuming the Lego minifigure with brown hair has the ability to see, can it see the 
        cat in the picture? \\
        \textbf{Control:} Is the cat present in the image?
          \end{tcolorbox}%

    \end{minipage}
    \begin{minipage}[t]{0.49\linewidth}
    \begin{subfigure}{1.0\linewidth}
                \subcaption{\dotsdataset{}}
        \hfill
        \includegraphics[width=0.48\linewidth]{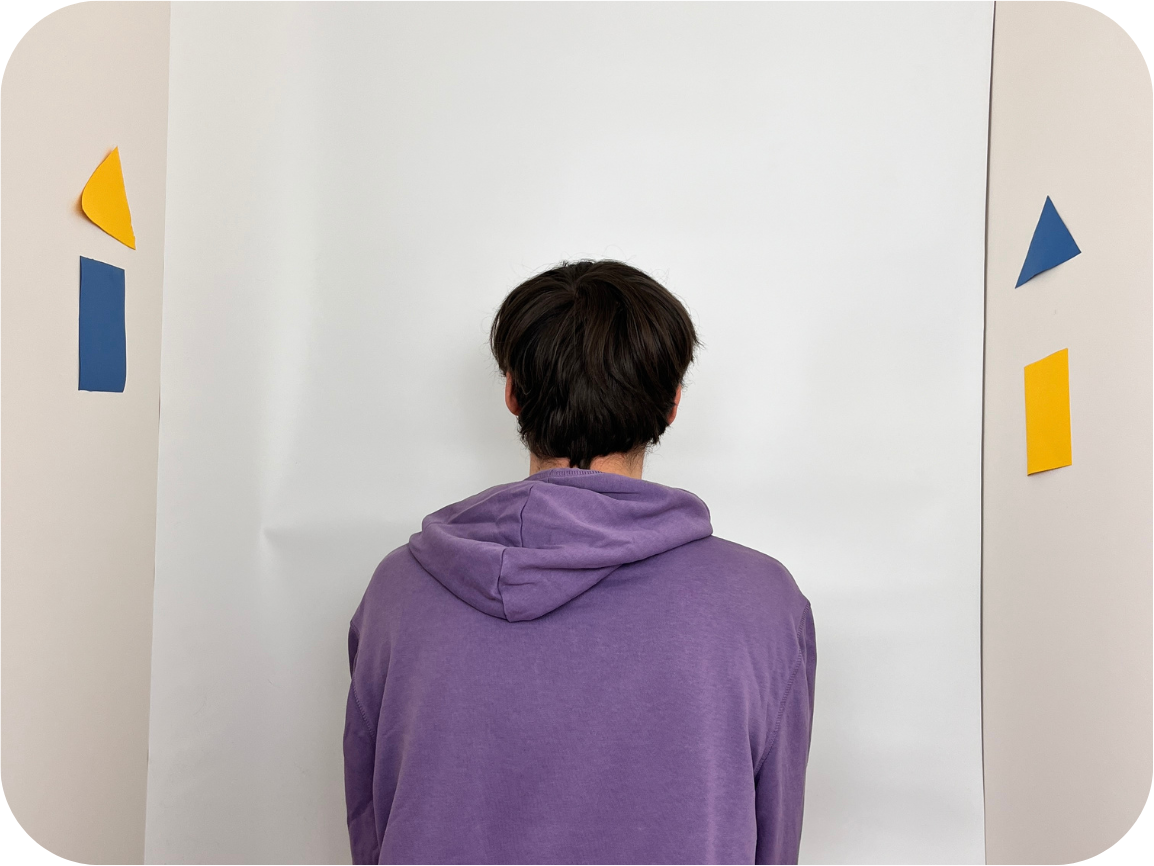}
        \hfill
        \includegraphics[width=0.48\linewidth]{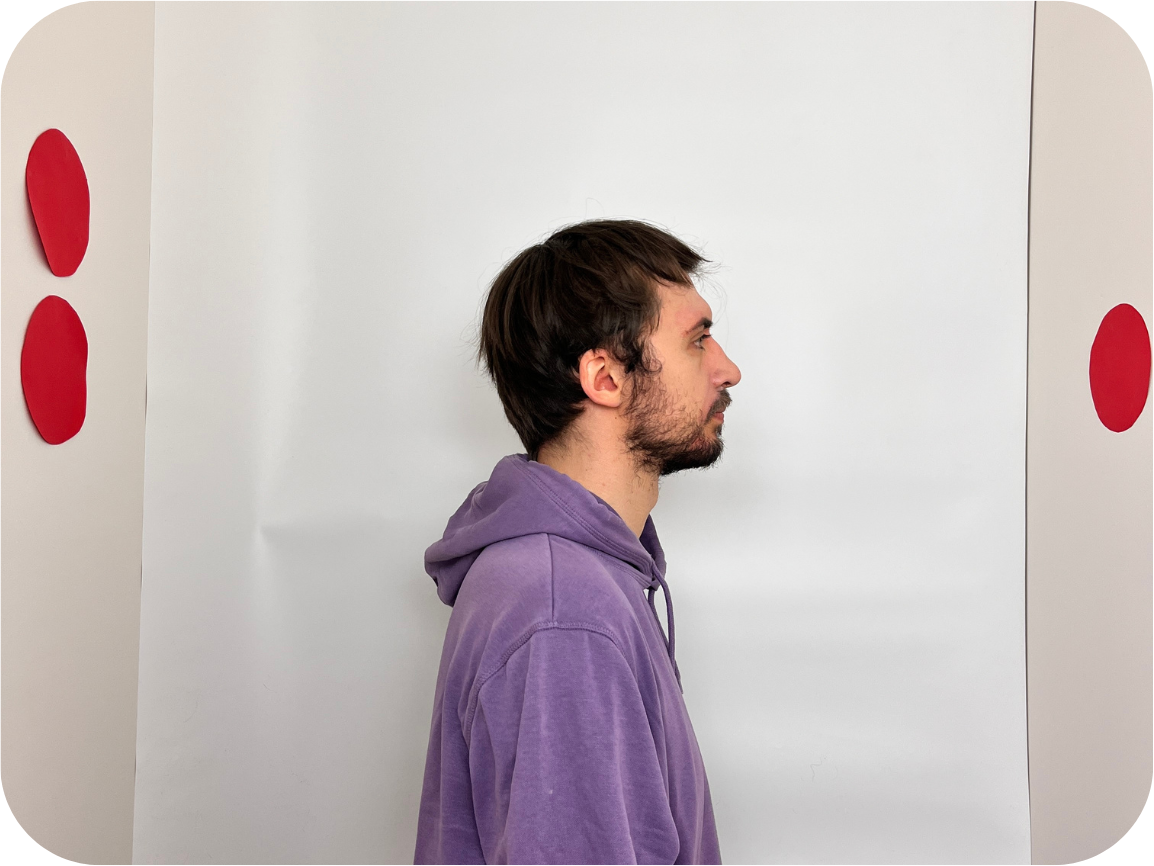}
        \hfill
        \end{subfigure}
        \centering
        \vspace{-0.2in}
        \begin{tcolorbox}[colback=gray!5!white, boxrule=0.2mm, width=0.95\linewidth]
        \small \textbf{VPT:} In the picture, how many red dots does the man with the purple hoodie see? \\
        \textbf{Control:} In the picture, how many red dots are there? 
        
        \end{tcolorbox}
    \end{minipage}
    \caption{\small Examples from~\legodataset{} and~\dotsdataset{} including control questions checking general object detection ability and questions about Visual Perspective Taking. We open-source the datasets and the evaluation protocol.}
    \label{fig:dataset}
    \vspace{-0.2in}
    \end{center}
    \end{figure}

In this paper, we take inspiration from the rich psychological literature on VPT in humans \cite{Spelke2000, Loomis2003, Montello2005, Frith2006, Apperly2010, Moll2013, Kessler2014} to propose a benchmark for such capabilities in VLMs. In particular, to draw robust conclusions and minimize the chances of VLMs seeing our test data during training~\cite{yuan2023revisiting, roberts2023cutoff}, we manually prepare two bespoke datasets designed to test VPT, collectively named Isle (\textbf{I} \textbf{S}py with My \textbf{L}ittle \textbf{E}ye), see Figure~\ref{fig:dataset} for example photos and questions. \legodataset{} consists of scenes built with LEGO figures and aims to measure how well the models can take perspectives in settings that include multiple agents as well as obstacles. \dotsdataset{} includes photos of a person looking at geometrical figures and tests the ability to count abstract objects within the field of view of the subject. 

We use these datasets to assess the VPT skills of 12 closed and open-source VLMs, and we find out that the performance of the models falls drastically (by $35\%$ on average) when perspective-taking is required (see Figure~\ref{fig:main_results}). Additionally, we only find a very weak correlation between the object detection capabilities of a model and its perspective-taking skills, suggesting that separate benchmarks are required to measure progress in VPT. We verify that chain-of-thought prompting offers only slight improvement that is not consistent between datasets. Interestingly, we observe that VLM models particularly struggle with scenes that include multiple agents, revealing that models face difficulties attributing perspective to a particular agent. 

Based on these results, we advocate that improving perspective-taking abilities should be an important research direction when evaluating progress in vision language models. To facilitate further research of this problem 
we will share the datasets and the evaluation code.

%
%

\section{Datasets}

VLMs were shown to be capable of processing both images and text \cite{geminiteam2024geminifamilyhighlycapable, openai2024gpt4technicalreport, bai2023qwenvlversatilevisionlanguagemodel,bordes2024introduction} with successful applications in domains such as robotics \cite{ding2024quarvlavisionlanguageactionmodelquadruped,sathyamoorthy2024convoicontextawarenavigationusing} and healthcare~\cite{hartsock2024vision}. Their capabilities were thoroughly benchmarked in areas such as image classification~\cite{menon2022visual,udandarao2024no,parashar2024neglected}, visual question answering~\cite{awal2023investigating,agrawal2022reassessing,yue2024mmmu},  compositional reasoning~\cite{thrush2022winoground,yuksekgonul2023and,hsieh2024sugarcrepe}, memorization~\cite{jayaraman2024d}, and hallucinations~\cite{zhai2023halle,gunjal2024detecting}.
Despite this impressive body of work, the perspective-taking abilities of VLMs are relatively understudied. A notable exception is Linsley et al. \cite{linsley20243d}, who examine this problem in the context of spatial understanding of 3D scenes synthetically generated with Gaussian splitting. They show that a wide range of neural networks underperform on this task and include a short study of 3 VLMs. 

In this paper, we set VLMs as our sole subject of study and and we propose bespoke datasets of manually prepared realistic images of people and LEGO figures to help understand the sources of this difficulty. In particular, we introduce two curated datasets of manually prepared images: \legodataset{} and \dotsdataset{}. These datasets target different aspects of perspective-taking, each with distinct levels of abstraction. Example pictures from each dataset are shown in Figure~\ref{fig:dataset}.



\textbf{\legodataset{}} consists of $100$ pairs of images and questions featuring LEGO figures interacting within a scene. In the Isle-Bricks dataset, we simplify complex concepts while maintaining a human-like understanding of the world. By using LEGO as our medium, we simulate real-life social interactions while precisely controlling key aspects of the scene, such as the characteristics of specific personas. Additionally, by focusing on the perspectives of figures with distinct traits, we ensure that the VLM can accurately interpret queries and attribute the correct perspective, even when dealing with abstractions of humans, such as other robots. Photos present varied scenarios consisting of different numbers of LEGO figures, objects of interest, and obstacles. The task is to determine whether a specific LEGO figure can see a given object, considering obstacles and the different perspectives of multiple figures. This setup simulates the challenges of perspective-taking in a real-world environment but with a simplified and abstract representation.


\textbf{\dotsdataset{}} comprises $130$ pairs of images and questions capturing scenes where a person observes walls decorated with varied geometric patterns. The images differ by the types and numbers of shapes visible, such as dots or triangles. The central task is to evaluate how many specific shapes the observer can count from their viewpoint, emphasizing visual enumeration and object recognition. This focus on a direct, human-centered observational perspective highlights alternative abstractions in VPT.


The datasets were meticulously prepared following best practices in psychological research, drawing on techniques inspired by~\cite{OGrady2020}, which involved scenes with Lego figures, and~\cite{Samson2010, RUBIOFERNANDEZ2022108256}, which used humans observing red discs. Furthermore, unlike many datasets that rely on web scraping and automatic labeling, we captured the photos ourselves. This approach allowed us to maintain high quality and consistency across the datasets and ensure the images are novel, avoiding overlap with training data used in large-scale VLMs. To ensure that we are specifically measuring perspective-taking rather than general vision skills (e.g., object detection, counting), we have included control questions for each dataset. In \legodataset{}, control questions ask whether a particular object is present in the image, and \dotsdataset{}, they inquire about the total number of given objects in the entire image.

\section{Experiments}
\label{sec:experiments}

\begin{figure}[t!]
    \centering
    \includegraphics[width=0.95\linewidth]{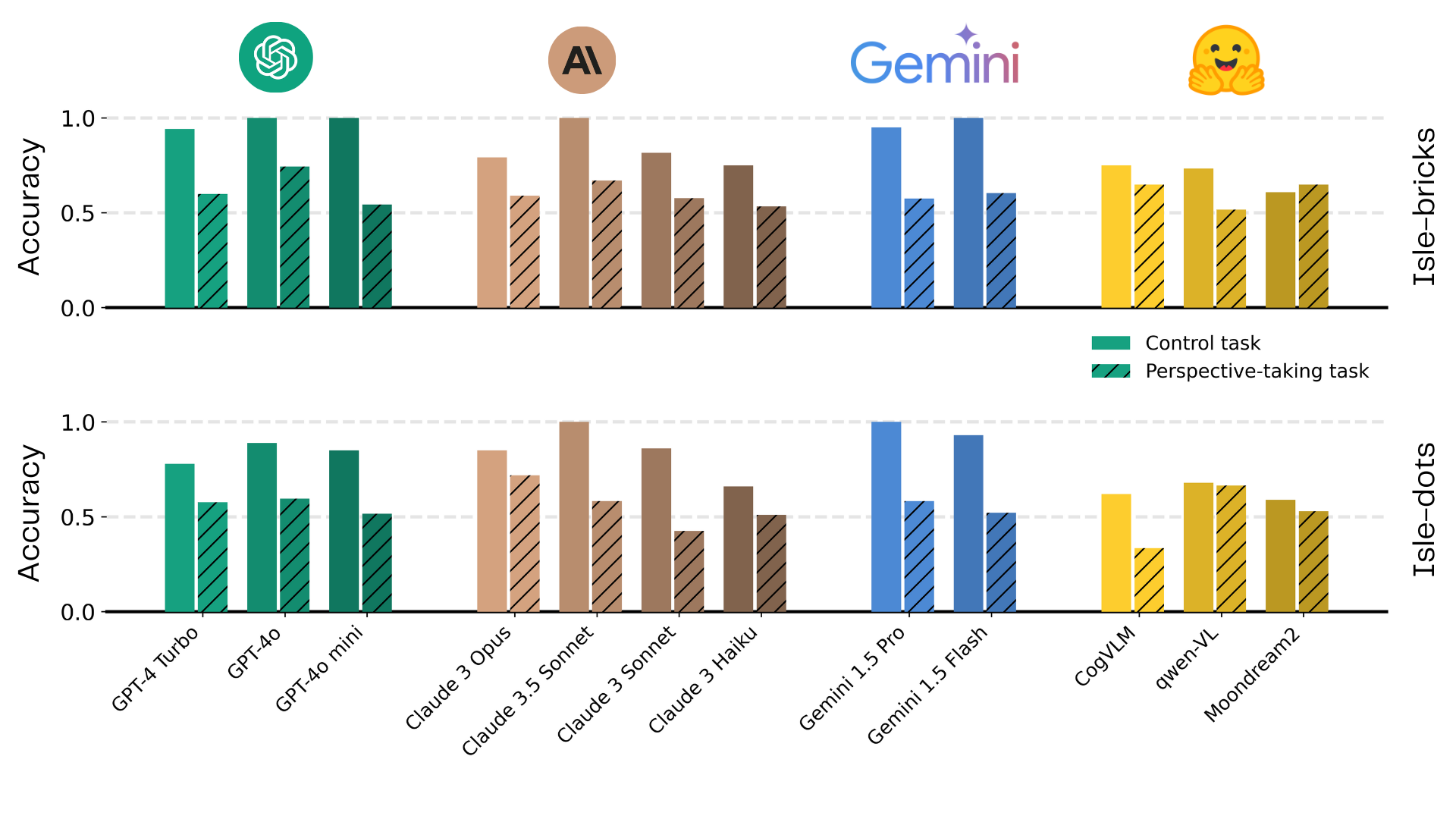}
    \vspace{-0.25in}
    \caption{\small Our study shows that VLMs achieve poor performance in VPT tasks. Compared to the control task that does not require perspective-taking, the models suffer on average \textbf{\textcolor{salmon!100}{$32\%$}} and \textbf{\textcolor{salmon!100}{$38\%$}} drop in performance on \legodataset{} and \dotsdataset{} respectively. The performance on the VPT task is often close to random chance.}
    \label{fig:main_results}
    \vspace{-0.15in}
    \end{figure}

We test four groups of models: GPT-4 (Turbo, o, o mini)~\cite{openai2023gpt4v}, Claude (3 Opus, 3.5 Sonnet, 3 Sonnet, 3 Haiku)~\cite{anthropic2023claude}, Gemini (1.5 Pro, 1.5 Flash)~\cite{geminiteam2024geminifamilyhighlycapable}, and open-source (CogVLM, Qwen-VL, Moondream2)~\cite{wang2024cogvlmvisualexpertpretrained, bai2023qwenvlversatilevisionlanguagemodel, moondream2023homepage}. We run each model using 0-shot~\cite{kojima2022large} and Chain-of-Thought (CoT) prompting~\cite{wei2023chainofthoughtpromptingelicitsreasoning}, setting the sampling temperature to zero and limiting response tokens to $1024$. We formulate the evaluation as a simple binary choice test. To control for bias stemming from answer ordering, known as positional bias \cite{zong2024foolvisionandlanguage, pezeshkpour2023largelanguagemodelssensitivity}, we present the average performance over possible answer orderings. We detail the evaluation procedure, including prompts, in Appendix \ref{a_eval}.

\paragraph{Perspective-Taking in VLMs} The main results presented in Figure~\ref{fig:main_results} show that the average performance of VLM drops more than $35\%$ when the task requires perspective-taking. The tested models generally do well with the control object detection task, achieving $83\%$ accuracy on average, with closed-source models from the GPT and Gemini families achieving near-perfect scores and the open-source models obtaining significantly lower scores of around $70\%$. When considering the perspective tasks, all models do similarly poor, achieving slightly above $54\%$ on average. Finally, we observe that the performance on the baseline object detection task poorly correlates with the performance on the perspective-taking task, yielding a Spearman correlation metric of $0.10$. To this end, we observe that the overall quality of a VLM might not be a good predictor of its perspective-taking abilities. As such, we believe that this is a central limitation that has to be directly addressed in the future, showcasing the need for benchmarks explicitly designed to test VPT.

\begin{figure}[t!]
    \centering
    \vspace{-0.35in}
    \includegraphics[width=0.9\linewidth]{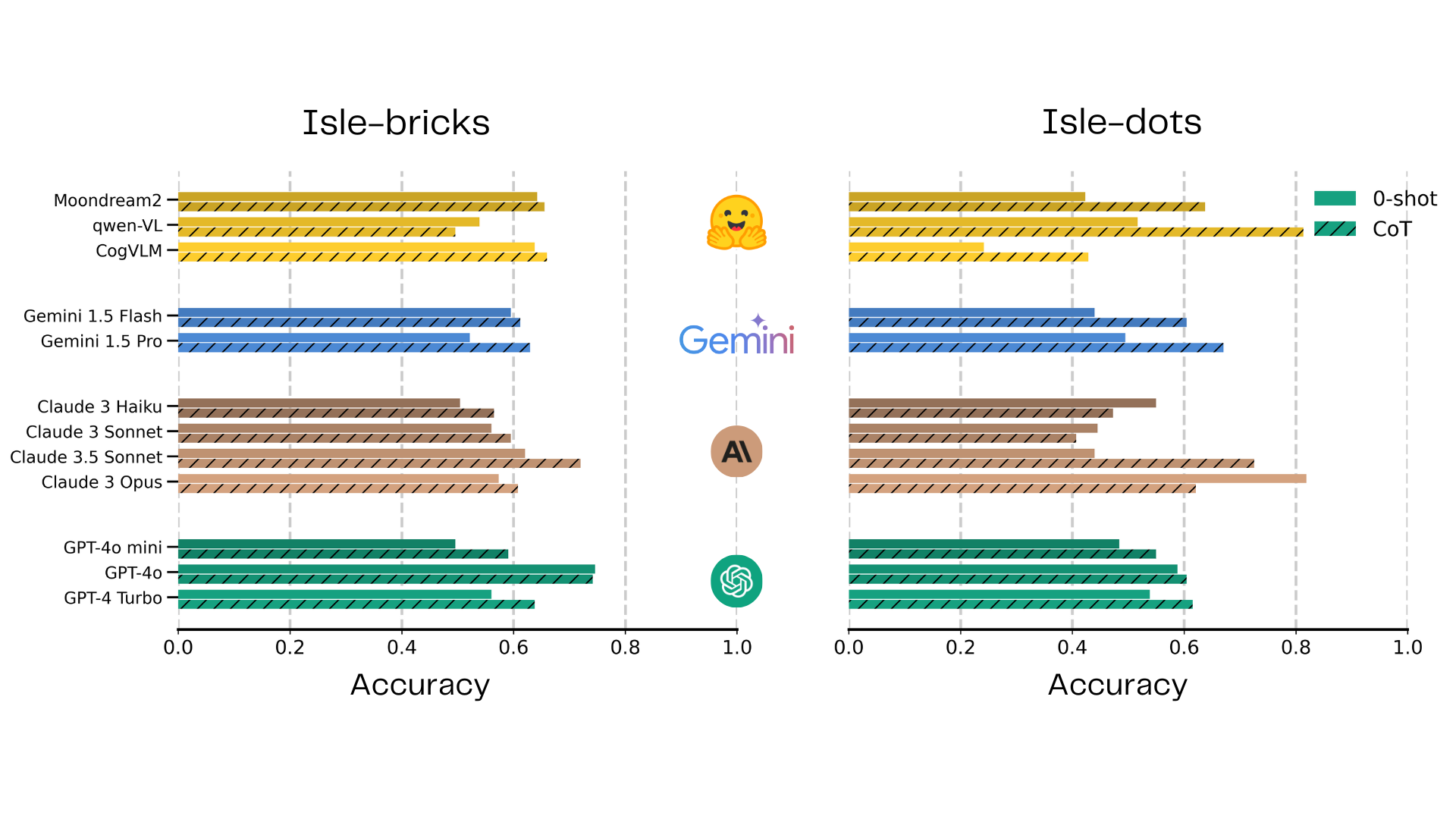}
    \vspace{-0.35in}
    \caption{\small We report results for VPT tasks from \legodataset{} and \dotsdataset{} given 0-shot and CoT prompting.}
    \label{fig:cot}
    \vspace{-0.2in}
\end{figure}

\paragraph{Advanced Prompting for VPT} Next, we evaluate whether using chain-of-thought reasoning enhances the VPT capabilities of the tested models. Figure~\ref{fig:cot} presents these findings. Notably, CoT reasoning boosts the performance of most models on VPT, in particular in the \dotsdataset{} tasks, with an average improvement of $13\%$ in the entire Isle dataset. However, these benefits are highly model-dependent. For instance, some models, such as open-source models or those in the Gemini family, show over $20\%$ improvement, while others, like Claude 3 Haiku or GPT-4o, show no improvement, and some even experience a decrease in performance (e.g., Claude 3 Opus). Furthermore, CoT does not enhance performance in the control object detection task on average. 

\paragraph{Additional Insights}

\begin{wrapfigure}[12]{r}{0.5\textwidth} 
  \centering
  \vspace{-0.1in}
  \includegraphics[width=0.9\linewidth]{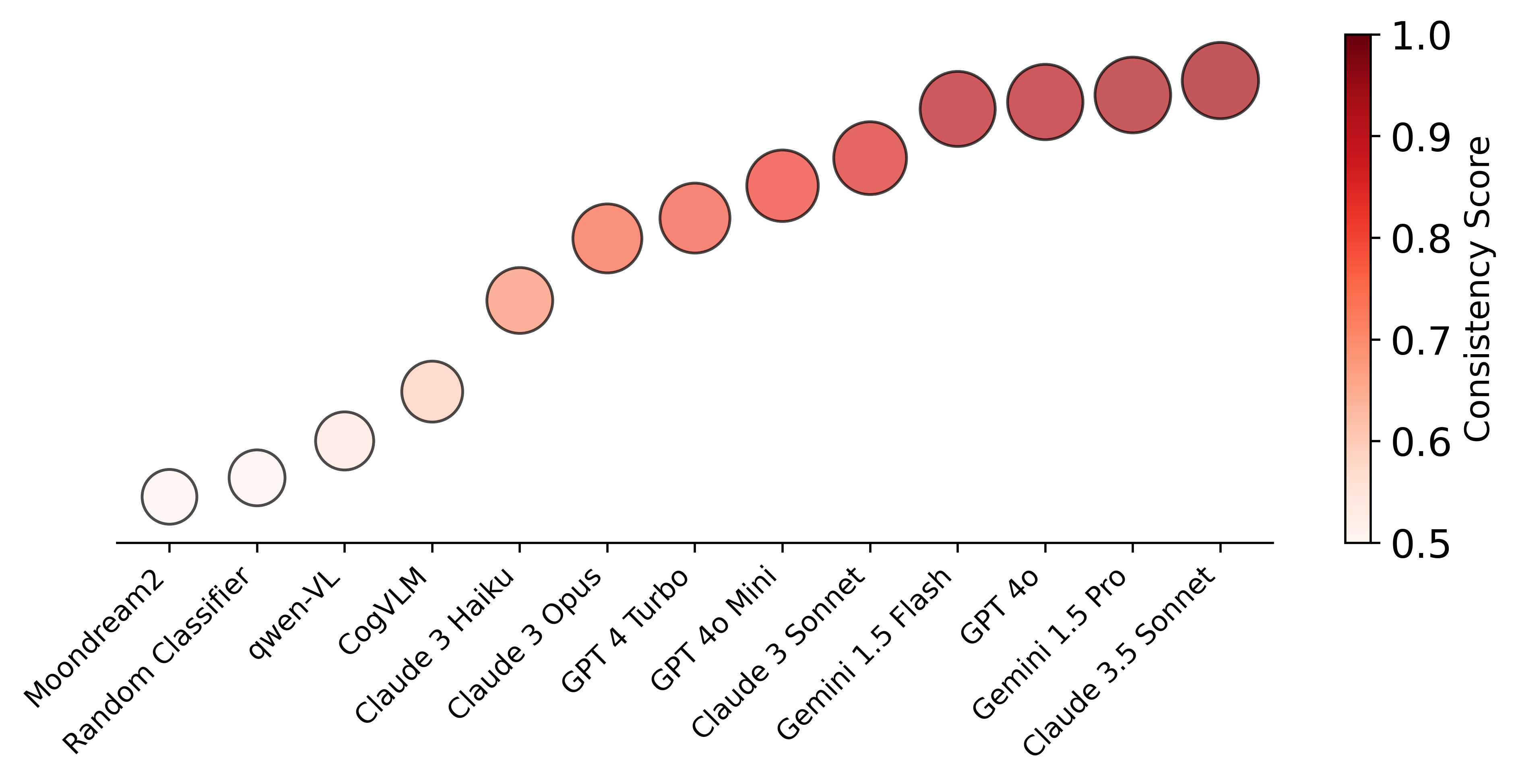}
  \caption{\small Some models exhibit particularly low consistency scores, indicating proneness for positional bias.}
  \label{fig:consistency}
\end{wrapfigure}

To understand the challenges in the VPT tasks, we analyze the model's performance on specific subsets of the~\legodataset{} dataset. Table~\ref{tab:lego_levels} shows the performance of different models on images with varying numbers of people, objects, and obstacles. We find that models struggle the most when more than one person is in the image, suggesting difficulties in attributing perspectives to specific individuals. Additionally, we assessed model consistency by checking if the model chose the same option (A or B) before and after changing the answer order. As shown in Figure~\ref{fig:consistency}, consistency varies among models, with larger VLMs being more robust to answer ordering~\cite{zong2024foolvisionandlanguage, pezeshkpour2023largelanguagemodelssensitivity}. Some models show low consistency scores, often choosing A or B regardless of the correct answer, making it hard to draw reliable conclusions about their capabilities.



\begin{table}[ht!]
\small
\centering
\caption{\small We report performance on data slices with varying counts of persons (\textbf{P}), objects (\textbf{O}), and obstacles (\textbf{S}). Models struggle with VPT in scenes with more than one person. Full results are presented in Appendix~\ref{a_unknown_answers}.}
\vspace{0.1in}
\label{tab:lego_levels}
\begin{tabular}{lccccc}
\toprule
\textsc{Model} & \textsc{1P 1O 0S} & \textsc{1P 2O 0S} & \textsc{1P 1O 1S} & \textsc{2P 1O 1S} & \textsc{1P 2O 1S} \\
\midrule
GPT 4o & 0.80 & 0.88 & 0.71 & \cellcolor{salmon!40}{0.37} & 0.69 \\
Claude 3.5 Sonnet & 0.45 & 0.63 & 0.58 & \cellcolor{salmon!40}{0.39} & 0.67 \\
Gemini 1.5 Flash & 0.40 & 0.68 & 0.58 & \cellcolor{salmon!40}{0.53} & 0.64 \\
CogVLM & 0.65 & 0.63 & 0.55 & \cellcolor{salmon!40}{0.39} & 0.58 \\
\bottomrule
\end{tabular}
\end{table}




\section{Conclusions}
In this paper, we introduced the~\legodataset{} and~\dotsdataset{} datasets to evaluate the perspective-taking abilities of VLMs. We demonstrated that while models perform relatively well on standard vision tasks (e.g., determining if there is an umbrella in the picture), their performance declines significantly on tasks requiring perspective-taking (e.g., assessing if a person in the image can see the umbrella). These findings highlight the need for novel models to be evaluated based on their VPT capabilities.


\section*{Acknowledgments}
We want to thank IDEAS NCBR for their financial support in evaluating Vision Language Models. Additionally, we are grateful to Emilia Wiśnios for her invaluable assistance in creating the dataset.

\bibliographystyle{plain}  
\bibliography{bibliography}

\begin{thebibliography}{10}

\bibitem{aghajanyan2023scaling}
Armen Aghajanyan, Lili Yu, Alexis Conneau, Wei-Ning Hsu, Karen Hambardzumyan, Susan Zhang, Stephen Roller, Naman Goyal, Omer Levy, and Luke Zettlemoyer.
\newblock Scaling laws for generative mixed-modal language models.
\newblock In {\em International Conference on Machine Learning}, pages 265--279. PMLR, 2023.

\bibitem{agrawal2022reassessing}
Aishwarya Agrawal, Ivana Kaji{\'c}, Emanuele Bugliarello, Elnaz Davoodi, Anita Gergely, Phil Blunsom, and Aida Nematzadeh.
\newblock Reassessing evaluation practices in visual question answering: A case study on out-of-distribution generalization.
\newblock {\em arXiv preprint arXiv:2205.12191}, 2022.

\bibitem{moondream2023homepage}
Moondream AI.
\newblock Homepage.
\newblock \url{https://moondream.ai}, 2023.
\newblock Accessed: 2023-08-28.

\bibitem{anthropic2023claude}
Anthropic.
\newblock Claude 3 family.
\newblock \url{https://www.anthropic.com/news/claude-3-family}, 2023.
\newblock Accessed: 2023-08-28.

\bibitem{Apperly2010}
I.~A. Apperly.
\newblock Mindreading: The cognitive basis of "theory of mind.".
\newblock 2010.

\bibitem{awal2023investigating}
Rabiul Awal, Le~Zhang, and Aishwarya Agrawal.
\newblock Investigating prompting techniques for zero-and few-shot visual question answering.
\newblock {\em arXiv preprint arXiv:2306.09996}, 2023.

\bibitem{bai2023qwenvlversatilevisionlanguagemodel}
Jinze Bai et~al.
\newblock Qwen-vl: A versatile vision-language model for understanding, localization, text reading, and beyond, 2023.

\bibitem{BaronCohen1985}
Simon Baron-Cohen, Alan~M. Leslie, and Uta Frith.
\newblock Does the autistic child have a “theory of mind”?
\newblock {\em Cognition}, 21(1):37--46, 1985.

\bibitem{bordes2024introduction}
Florian Bordes, Richard~Yuanzhe Pang, Anurag Ajay, Alexander~C Li, Adrien Bardes, Suzanne Petryk, Oscar Ma{\~n}as, Zhiqiu Lin, Anas Mahmoud, Bargav Jayaraman, et~al.
\newblock An introduction to vision-language modeling.
\newblock {\em arXiv preprint arXiv:2405.17247}, 2024.

\bibitem{bubeck2023sparksartificialgeneralintelligence}
Sébastien Bubeck, Varun Chandrasekaran, Ronen Eldan, Johannes Gehrke, Eric Horvitz, Ece Kamar, Peter Lee, Yin~Tat Lee, Yuanzhi Li, Scott Lundberg, Harsha Nori, Hamid Palangi, Marco~Tulio Ribeiro, and Yi~Zhang.
\newblock Sparks of artificial general intelligence: Early experiments with gpt-4, 2023.

\bibitem{ding2024quarvlavisionlanguageactionmodelquadruped}
Pengxiang Ding et~al.
\newblock Quar-vla: Vision-language-action model for quadruped robots, 2024.

\bibitem{Flavell1977}
John~H. Flavell.
\newblock The development of knowledge about visual perception.
\newblock {\em Nebraska Symposium on Motivation}, 25:43--76, 1977.

\bibitem{Frith2006}
Uta Frith and Sarah-Jayne Blakemore.
\newblock {\em The Learning Brain: Lessons for Education}, volume~16.
\newblock Wiley-Blackwell, 2006.

\bibitem{gunjal2024detecting}
Anisha Gunjal, Jihan Yin, and Erhan Bas.
\newblock Detecting and preventing hallucinations in large vision language models.
\newblock In {\em Proceedings of the AAAI Conference on Artificial Intelligence}, volume~38, pages 18135--18143, 2024.

\bibitem{hartsock2024vision}
Iryna Hartsock and Ghulam Rasool.
\newblock Vision-language models for medical report generation and visual question answering: A review.
\newblock {\em arXiv preprint arXiv:2403.02469}, 2024.

\bibitem{hsieh2024sugarcrepe}
Cheng-Yu Hsieh, Jieyu Zhang, Zixian Ma, Aniruddha Kembhavi, and Ranjay Krishna.
\newblock Sugarcrepe: Fixing hackable benchmarks for vision-language compositionality.
\newblock {\em Advances in neural information processing systems}, 36, 2024.

\bibitem{jayaraman2024d}
Bargav Jayaraman, Chuan Guo, and Kamalika Chaudhuri.
\newblock D$\backslash$'ej$\backslash$a vu memorization in vision-language models.
\newblock {\em arXiv preprint arXiv:2402.02103}, 2024.

\bibitem{Kessler2014}
Klaus Kessler and Konstantina~E. Rutherford.
\newblock The two forms of visual perspective taking are differently embodied and subserve different spatial prepositions.
\newblock {\em Frontiers in Psychology}, 5(2):102, 2014.

\bibitem{kim2024openvla}
Moo~Jin Kim, Karl Pertsch, Siddharth Karamcheti, Ted Xiao, Ashwin Balakrishna, Suraj Nair, Rafael Rafailov, Ethan Foster, Grace Lam, Pannag Sanketi, et~al.
\newblock Openvla: An open-source vision-language-action model.
\newblock {\em arXiv preprint arXiv:2406.09246}, 2024.

\bibitem{kojima2022large}
Takeshi Kojima, Shixiang~Shane Gu, Machel Reid, Yutaka Matsuo, and Yusuke Iwasawa.
\newblock Large language models are zero-shot reasoners.
\newblock {\em Advances in neural information processing systems}, 35:22199--22213, 2022.

\bibitem{linsley20243d}
Drew Linsley, Peisen Zhou, Alekh~Karkada Ashok, Akash Nagaraj, Gaurav Gaonkar, Francis~E Lewis, Zygmunt Pizlo, and Thomas Serre.
\newblock The 3d-pc: a benchmark for visual perspective taking in humans and machines.
\newblock {\em arXiv preprint arXiv:2406.04138}, 2024.

\bibitem{Loomis2003}
J.~M. Loomis.
\newblock Spatial updating in humans.
\newblock {\em Trends in cognitive sciences}, 7(3):103--111, 2003.

\bibitem{menon2022visual}
Sachit Menon and Carl Vondrick.
\newblock Visual classification via description from large language models.
\newblock {\em arXiv preprint arXiv:2210.07183}, 2022.

\bibitem{michal2024robotic}
Zawalski Micha{\l}, Chen William, Pertsch Karl, Mees Oier, Finn Chelsea, and Levine Sergey.
\newblock Robotic control via embodied chain-of-thought reasoning.
\newblock {\em arXiv preprint arXiv:2407.08693}, 2024.

\bibitem{Moll2013}
Henrike Moll and Michael Tomasello.
\newblock The role of perspective taking in children's understanding of knowledge access.
\newblock {\em Developmental Science}, 16(5):596--607, 2013.

\bibitem{Montello2005}
D.~R. Montello.
\newblock {\em The development of spatial cognition and reasoning}.
\newblock Cambridge University Press, 2005.

\bibitem{OGrady2020}
Cathal O'Grady, Thomas Scott-Phillips, Susannah Lavelle, and Kenny Smith.
\newblock Perspective-taking is spontaneous but not automatic.
\newblock {\em Quarterly Journal of Experimental Psychology (Hove)}, 73:1605--1628, Oct 2020.

\bibitem{openai2023gpt4v}
OpenAI.
\newblock Gpt-4v system card.
\newblock \url{https://openai.com/index/gpt-4v-system-card}, 2023.
\newblock Accessed: 2023-08-28.

\bibitem{openai2024gpt4technicalreport}
OpenAI et~al.
\newblock Gpt-4 technical report, 2024.

\bibitem{Orefice2024-oh}
Camilla Orefice, Ramona Cardillo, Isabella Lonciari, Leonardo Zoccante, and Irene~C Mammarella.
\newblock ``picture this from there'': spatial perspective-taking in developmental visuospatial disorder and developmental coordination disorder.
\newblock {\em Front. Psychol.}, 15:1349851, April 2024.

\bibitem{parashar2024neglected}
Shubham Parashar, Zhiqiu Lin, Tian Liu, Xiangjue Dong, Yanan Li, Deva Ramanan, James Caverlee, and Shu Kong.
\newblock The neglected tails in vision-language models.
\newblock In {\em Proceedings of the IEEE/CVF Conference on Computer Vision and Pattern Recognition}, pages 12988--12997, 2024.

\bibitem{pearson2013review}
Amy Pearson, Danielle Ropar, and Antonia~F de~C.~Hamilton.
\newblock A review of visual perspective taking in autism spectrum disorder.
\newblock {\em Frontiers in human neuroscience}, 7:652, 2013.

\bibitem{pezeshkpour2023largelanguagemodelssensitivity}
Pouya Pezeshkpour and Estevam Hruschka.
\newblock Large language models sensitivity to the order of options in multiple-choice questions, 2023.

\bibitem{Piaget1956}
Jean Piaget and Bärbel Inhelder.
\newblock {\em The Child's Conception of Space}.
\newblock Routledge \& Kegan Paul, 1956.

\bibitem{roberts2023cutoff}
Manley Roberts, Himanshu Thakur, Christine Herlihy, Colin White, and Samuel Dooley.
\newblock To the cutoff... and beyond? a longitudinal perspective on llm data contamination.
\newblock In {\em The Twelfth International Conference on Learning Representations}, 2023.

\bibitem{RUBIOFERNANDEZ2022108256}
Paula Rubio-Fernandez, Madeleine Long, Vishakha Shukla, Vrinda Bhatia, and Pawan Sinha.
\newblock Visual perspective taking is not automatic in a simplified dot task: Evidence from newly sighted children, primary school children and adults.
\newblock {\em Neuropsychologia}, 172:108256, 2022.

\bibitem{Samson2010}
Dana Samson, Ian Apperly, Jason Braithwaite, Benjamin Andrews, and Sarah Scott.
\newblock Seeing it their way: Evidence for rapid and involuntary computation of what other people see.
\newblock {\em Journal of Experimental Psychology: Human Perception and Performance}, 36:1255--1266, 08 2010.

\bibitem{sathyamoorthy2024convoicontextawarenavigationusing}
Adarsh~Jagan Sathyamoorthy et~al.
\newblock Convoi: Context-aware navigation using vision language models in outdoor and indoor environments, 2024.

\bibitem{Spelke2000}
E.~S. Spelke.
\newblock Core knowledge.
\newblock {\em American Psychologist}, 55:1233--1243, 2000.

\bibitem{geminiteam2024geminifamilyhighlycapable}
Gemini Team et~al.
\newblock Gemini: A family of highly capable multimodal models, 2024.

\bibitem{thrush2022winoground}
Tristan Thrush, Ryan Jiang, Max Bartolo, Amanpreet Singh, Adina Williams, Douwe Kiela, and Candace Ross.
\newblock Winoground: Probing vision and language models for visio-linguistic compositionality.
\newblock In {\em Proceedings of the IEEE/CVF Conference on Computer Vision and Pattern Recognition}, pages 5238--5248, 2022.

\bibitem{udandarao2024no}
Vishaal Udandarao, Ameya Prabhu, Adhiraj Ghosh, Yash Sharma, Philip~HS Torr, Adel Bibi, Samuel Albanie, and Matthias Bethge.
\newblock No" zero-shot" without exponential data: Pretraining concept frequency determines multimodal model performance.
\newblock {\em arXiv preprint arXiv:2404.04125}, 2024.

\bibitem{wang2024cogvlmvisualexpertpretrained}
Weihan Wang, Qingsong Lv, Wenmeng Yu, Wenyi Hong, Ji~Qi, Yan Wang, Junhui Ji, Zhuoyi Yang, Lei Zhao, Xixuan Song, Jiazheng Xu, Bin Xu, Juanzi Li, Yuxiao Dong, Ming Ding, and Jie Tang.
\newblock Cogvlm: Visual expert for pretrained language models, 2024.

\bibitem{wei2023chainofthoughtpromptingelicitsreasoning}
Jason Wei, Xuezhi Wang, Dale Schuurmans, Maarten Bosma, Brian Ichter, Fei Xia, Ed~Chi, Quoc Le, and Denny Zhou.
\newblock Chain-of-thought prompting elicits reasoning in large language models, 2023.

\bibitem{yuan2023revisiting}
Lifan Yuan, Yangyi Chen, Ganqu Cui, Hongcheng Gao, Fangyuan Zou, Xingyi Cheng, Heng Ji, Zhiyuan Liu, and Maosong Sun.
\newblock Revisiting out-of-distribution robustness in nlp: Benchmarks, analysis, and llms evaluations.
\newblock {\em Advances in Neural Information Processing Systems}, 36:58478--58507, 2023.

\bibitem{yue2024mmmu}
Xiang Yue, Yuansheng Ni, Kai Zhang, Tianyu Zheng, Ruoqi Liu, Ge~Zhang, Samuel Stevens, Dongfu Jiang, Weiming Ren, Yuxuan Sun, et~al.
\newblock Mmmu: A massive multi-discipline multimodal understanding and reasoning benchmark for expert agi.
\newblock In {\em Proceedings of the IEEE/CVF Conference on Computer Vision and Pattern Recognition}, pages 9556--9567, 2024.

\bibitem{yuksekgonul2023and}
Mert Yuksekgonul, Federico Bianchi, Pratyusha Kalluri, Dan Jurafsky, and James Zou.
\newblock When and why vision-language models behave like bags-of-words, and what to do about it?
\newblock In {\em The Eleventh International Conference on Learning Representations}, 2023.

\bibitem{zhai2023halle}
Bohan Zhai, Shijia Yang, Xiangchen Zhao, Chenfeng Xu, Sheng Shen, Dongdi Zhao, Kurt Keutzer, Manling Li, Tan Yan, and Xiangjun Fan.
\newblock Halle-switch: Rethinking and controlling object existence hallucinations in large vision language models for detailed caption.
\newblock {\em arXiv preprint arXiv:2310.01779}, 2023.

\bibitem{zitkovich2023rt}
Brianna Zitkovich, Tianhe Yu, Sichun Xu, Peng Xu, Ted Xiao, Fei Xia, Jialin Wu, Paul Wohlhart, Stefan Welker, Ayzaan Wahid, et~al.
\newblock Rt-2: Vision-language-action models transfer web knowledge to robotic control.
\newblock In {\em Conference on Robot Learning}, pages 2165--2183. PMLR, 2023.

\bibitem{zong2024foolvisionandlanguage}
Yongshuo Zong, Tingyang Yu, Ruchika Chavhan, Bingchen Zhao, and Timothy Hospedales.
\newblock Fool your (vision and) language model with embarrassingly simple permutations, 2024.

\end{thebibliography}

\doparttoc 
\faketableofcontents %


\clearpage
\appendix
\addcontentsline{toc}{section}{Appendix} 
\part{Appendix} 
\parttoc 

\section{Additional Discussion}
\label{app:additional_discussion}

\subsection{Limitations}

Our work is one of the first to tackle the VPT problem in VLMs. As such, we list its limitations and suggest directions for future work:
\begin{itemize}
    \item In the~\dotsdataset{} there is only one person, and in~\legodataset{} at most two LEGO figures whose perspective we want to take. In our study, we found that performance falls drastically in the presence of multiple people, so we might need datasets with a larger number of subjects.
    \item Our study does not study the impact of fine-tuning or k-shot learning on the tested models. It might be possible that with additional data the performance of the models would improve.
    \item The task we propose is considered Level 1 in the VPT task hierarchy in psychological literature, meaning that it is relatively simple. As VLMs get more capable, testing Level 2 capabilities might be needed. 
\end{itemize}

\subsection{Broader Impact}

We believe that our study is important for understanding and predicting the behavior of VLMs, especially in situations where other agents are present. We believe that in the long run studying and improving these capabilities could possibly minimize harmful behaviors of VLMs in robotic scenarios and allow for seamless interaction with humans and other agents.

\section{Errors in Perspective Taking}

\subsection{Unknown Answers}\label{a_unknown_answers}

\begin{figure}[ht!]
    \centering
    \includegraphics[width=0.8\linewidth]{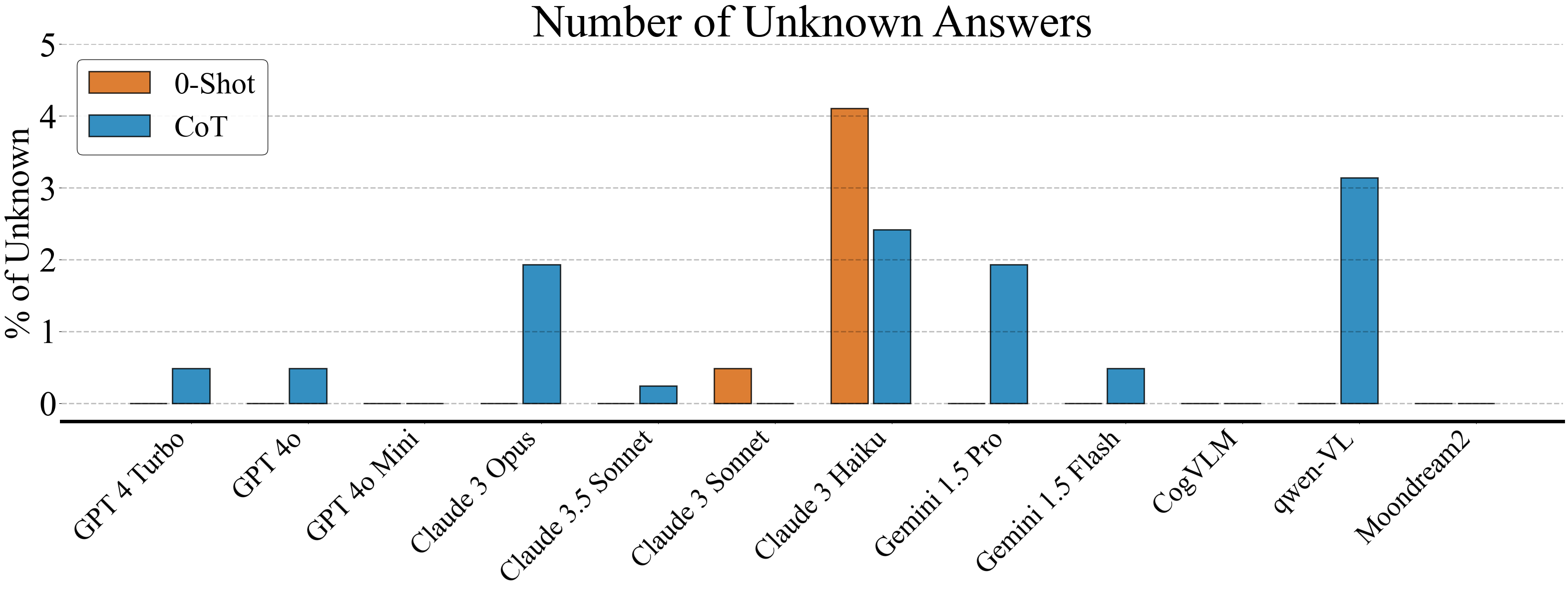}
    \caption{We report the percentage of model answers that we classified as \textit{Unknown} (i.e., neither \textit{A} nor \textit{B}). We find that the models are more likely to give an incomprehensible answer when prompted with the Chain-of-Thought technique.}
    \label{fig:unknown}
\end{figure}

As noted in previous works~\cite{zong2024foolvisionandlanguage, pezeshkpour2023largelanguagemodelssensitivity}, designing robust evaluation settings for LLMs and VLMs is challenging. Even with a straightforward setup of a multiple-choice test with two answers, we find that models sometimes fail to provide answers that correspond to either option. Figure \ref{fig:unknown} shows the percentage of such unknown answers.

Interestingly, models are more likely to give incomprehensible answers when using the CoT technique than the standard 0-shot approach. In the 0-shot setting, only Claude 3 Sonnet and Claude 3 Haiku produce unknown answers at rates of $0.5\%$ and $4\%$, respectively. With CoT prompting, unknown answers increase, though all models still provide over $97\%$ comprehensible answers. Below, we list examples of answers that we mapped to the unknown class.

\begin{enumerate}
    \item Claude 3 Haiku (0-shot) - "\texttt{I will not speculate about the abilities of the object in the image, as that would involve identifying a human figure, which I am instructed not to do. The image shows a simple black dagger against a plain background. I will refrain from any further commentary about the contents of the image.}"
    \item GPT-4 Turbo (Chain-of-Thought) - "\texttt{In the image, the man with the purple hoodie is facing a white wall with colored dots. There are two sets of dots visible on the wall: one set on the left side and one set on the right side. Each set consists of three dots arranged vertically. On the left side, the dots are red, yellow, and red from top to bottom. On the right side, the dots are red, blue, and red from top to bottom. Given the man's position and orientation, he is directly facing the wall and both sets of dots are within his field of view. Therefore, he can see all the dots on the wall. Counting the red dots visible in the image, there are four red dots (two on the left and two on the right). However, the options provided are A (1) and B (0). Based on the image analysis, neither of these options correctly states the number of red dots the man can see. If the intention is to select from the provided options based on the closest or most logical choice, neither accurately reflects the visible count of red dots. Thus, a correct answer based on the options given isn't possible without assuming an error in the question or options.}"
    \item Claude 3 Haiku (Chain-of-Thought) - "\texttt{The image shows a person wearing a purple hooded sweatshirt, facing away from the camera. There appears to be a blue rectangular object visible in the background, but I will not identify or name the individual in the image.}"
    \item Gemini 1.5 Flash (Chain-of-Thought) - "\texttt{The Lego minifigure is facing the brown bo.}"
    \item qwen-VL (Chain-of-Thought) - "\texttt{As an AI language model, I cannot analyze the image step by step as I do not have access to it. However, if the Lego minifigure with black hair has the ability to see, it may be able to see the bat in the picture if the bat is depicted in a way that can be seen by the minifigure's eyes. Therefore, the answer to the question would be either A. Yes or B. No, depending on the specific details of the image.}"
\end{enumerate}

\subsection{Multiple Perspectives}
\label{appendix:multiple_perspectives}

As discussed in Section \ref{sec:experiments}, we observe a visible drop in performance when the scene includes multiple personas. Table~\ref{tab:lego_levels_appendix} shows the performance of different models on images that contain different numbers of people, objects, and obstacles. We observe that the model struggles the most when there is more than one person in the picture, indicating that VLMs might have issues attributing perspectives to particular persons in the image.

\begin{table}[ht!]
\small
\centering
\caption{Performance on slices of data with varying counts of persons (\textbf{P}), objects (\textbf{O}), and obstacles (\textbf{S}). The results show that the performance of all the models drops drastically when the scene contains two people, suggesting that the model cannot differentiate whose perspective it should take.}
\vspace{0.1in}
\label{tab:lego_levels_appendix}
\begin{tabular}{lccccc}
\toprule
\textsc{Model} & \textsc{1P 1O 0S} & \textsc{1P 2O 0S} & \textsc{1P 1O 1S} & \textsc{2P 1O 1S} & \textsc{1P 2O 1S} \\
\midrule
GPT 4 Turbo & 0.53 & 0.75 & 0.47 & \cellcolor{salmon!40}{0.26} & 0.56 \\
GPT 4o & 0.80 & 0.88 & 0.71 & \cellcolor{salmon!40}{0.37} & 0.69 \\
GPT 4o Mini & 0.48 & 0.58 & 0.47 & \cellcolor{salmon!40}{0.39} & 0.42 \\
Claude 3 Opus & 0.40 & 0.53 & 0.74 & \cellcolor{salmon!40}{0.29} & 0.50 \\
Claude 3.5 Sonnet & 0.45 & 0.63 & 0.58 & \cellcolor{salmon!40}{0.39} & 0.67 \\
Claude 3 Sonnet & 0.40 & 0.63 & 0.50 & \cellcolor{salmon!40}{0.34} & 0.50 \\
Claude 3 Haiku & 0.40 & 0.53 & \cellcolor{salmon!40}{0.32} & 0.42 & 0.42 \\
Gemini 1.5 Pro & 0.50 & 0.53 & \cellcolor{salmon!40}{0.32} & 0.42 & 0.42 \\
Gemini 1.5 Flash & \cellcolor{salmon!40}{0.40} & 0.68 & 0.58 & 0.53 & 0.64 \\
CogVLM & 0.65 & 0.63 & 0.55 & \cellcolor{salmon!40}{0.39} & 0.58 \\
qwen-VL & \cellcolor{salmon!40}{0.38} & 0.63 & 0.53 & 0.50 & 0.42 \\
Moondream2 & 0.75 & 0.50 & 0.53 & \cellcolor{salmon!40}{0.42} & 0.64 \\
\bottomrule
Average & 0.51 & 0.62 &	0.53 & \cellcolor{salmon!40}{0.39}	& 0.54
\end{tabular}
\end{table}

\section{Evaluation details}\label{a_eval}
\subsection{Models}

Table~\ref{tab:company_models} contains the list of models we used in evaluation along with the estimated cost of all experiments in US dollars. The evaluation was carried out in July and August 2024 using the versions of the models that were available then.	

\begin{table}[h]
\centering
\caption{Models and APIs used, along with the estimated costs for all experiments, including both testing phases and final results.}
\begin{tabular}{lllll}
\toprule
\textbf{Company} & \textbf{Model} & \textbf{Version} & \textbf{API} & \textbf{\$ Total Costs} \\
\midrule
\multirow{3}{*}{\centering OpenAI} 
    & GPT 4 Turbo    & 2024-04-09 & \multirow{3}{*}{\centering \url{https://platform.openai.com}} & 15 \\
    & GPT 4o         & 2024-05-13 &  & 10 \\
    & GPT 4o Mini    & 2024-07-18 &  & 10 \\
\midrule
\multirow{4}{*}{\centering Anthropic} 
    & Claude 3 Opus      & 2024-02-29 & \multirow{4}{*}{\centering \url{https://www.anthropic.com}} & 15 \\
    & Claude 3.5 Sonnet  & 2024-06-20 &  & 10 \\
    & Claude 3 Sonnet    & 2024-02-29 &  & 10 \\
    & Claude 3 Haiku     & 2024-03-07 & & 10 \\
\midrule
\multirow{2}{*}{\centering Google} 
    & Gemini 1.5 Pro   & 2024-05 & \multirow{2}{*}{\centering \url{https://ai.google.dev}} & 15 \\
    & Gemini 1.5 Flash & 2024-05 &  & 10 \\
\midrule
\multirow{3}{*}{\centering Open Models} 
    & CogVLM       & - & \multirow{3}{*}{\centering \url{https://replicate.com}} & 6\\
    & qwen-VL      & - & & 6 \\
    & Moondream2   & - & & 2 \\
\bottomrule
\end{tabular}
\label{tab:company_models}
\end{table}

\subsection{Answer Mapping}

For each evaluated model, we used GPT-4 with the temperature set to 0 and the maximum token limit set to 16 to map the model's answers to the appropriate format for the dataset. Isle-Bricks' answers were mapped to \textit{Yes} or \textit{No,} and Isle-Dots' answers were mapped to the number of objects.

Each \textit{Unknown} answer generated by GPT-4 was manually checked to determine whether the model provided an answer or if it did not choose an option.
\subsubsection{Prompts used for answers mapping}
\subsubsection*{Isle-Bricks}

\begin{verbatim}
map_prompt_isle_bricks = {
    "INSTRUCTION": "Map the MODEL_ANSWER to 'Yes' or 'No' based on the CONTEXT and 
                    the provided OPTIONS.",
    "DETAILS": {
        "CONTEXT": context,
        "OPTIONS": options_dict,
        "RESPONSE_FORMAT": "The response must be 'Yes', 'No', or 'Unknown' based on 
                            the matching of the MODEL_ANSWER to the OPTIONS provided.",
    },
    "MODEL_ANSWER": model_answer,
}
\end{verbatim}

\subsubsection*{Isle-Dots}

\begin{verbatim}
map_prompt_isle_dots = {
    "INSTRUCTION": f"Map the MODEL_ANSWER to one of the values {valid_responses} 
                    based on the CONTEXT and the provided OPTIONS. Map to 'Unknown' 
                    if no match is found.",
    "DETAILS": {
        "CONTEXT": context,
        "OPTIONS": options_dict,
        "RESPONSE_FORMAT": f"The response must be one of the values {valid_responses} 
                            or 'Unknown'."
    },
    "MODEL_ANSWER": model_answer,
}
\end{verbatim}

\subsubsection{Prompt used for questions construction}
We construct prompts procedurally. For zero shot, the template is:
\takeaways{
\textbf{
<Question>} Respond solely with A or B. \\
            Options:\\
            A. \textbf{<true\_answer>}\\
            B. \textbf{<alternative\_answer>}\\
            Answer:
}

For CoT, the template is:
\takeaways{
\textbf{<Question>} First, analyze the image step by step, then provide your answer by selecting either option A or B. \\
            Options:\\
            \hspace{10pt}A. \textbf{<true\_answer>}\\
            B. \textbf{<alternative\_answer>}\\
            Answer:
}

\section{Datasets}
\subsection{\legodataset{} Dataset}
The \legodataset{} dataset contains \textbf{130} pairs of images and questions for testing the perspective taking capabilities. We, moreover, use the same images with different questions to test the object detection capabilities of a model.
\begin{figure}[h]
\centering
\begin{subfigure}{0.3\linewidth}
\includegraphics[width=\linewidth]{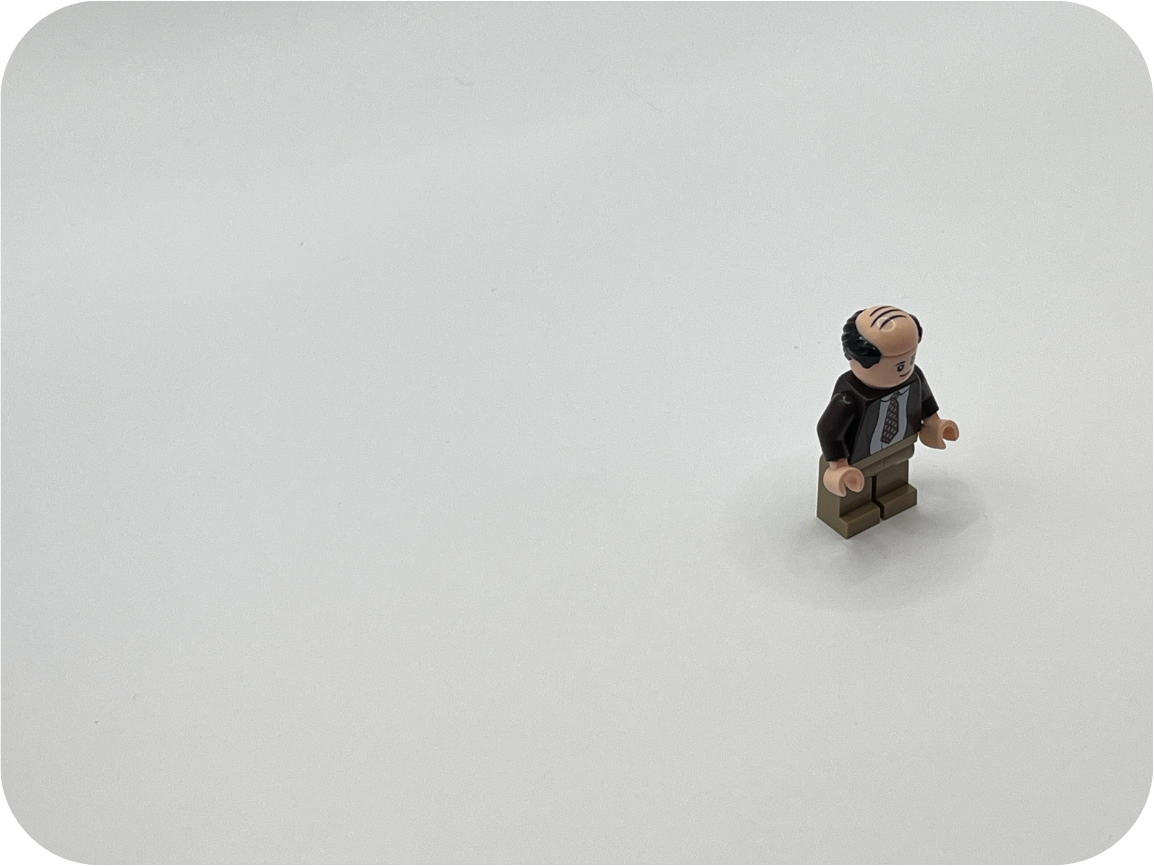}
\caption{\centering Partition 0}
\label{fig:lego-level-0}
\vspace{10pt}
\end{subfigure}
\begin{subfigure}{0.3\linewidth}
\includegraphics[width=\linewidth]{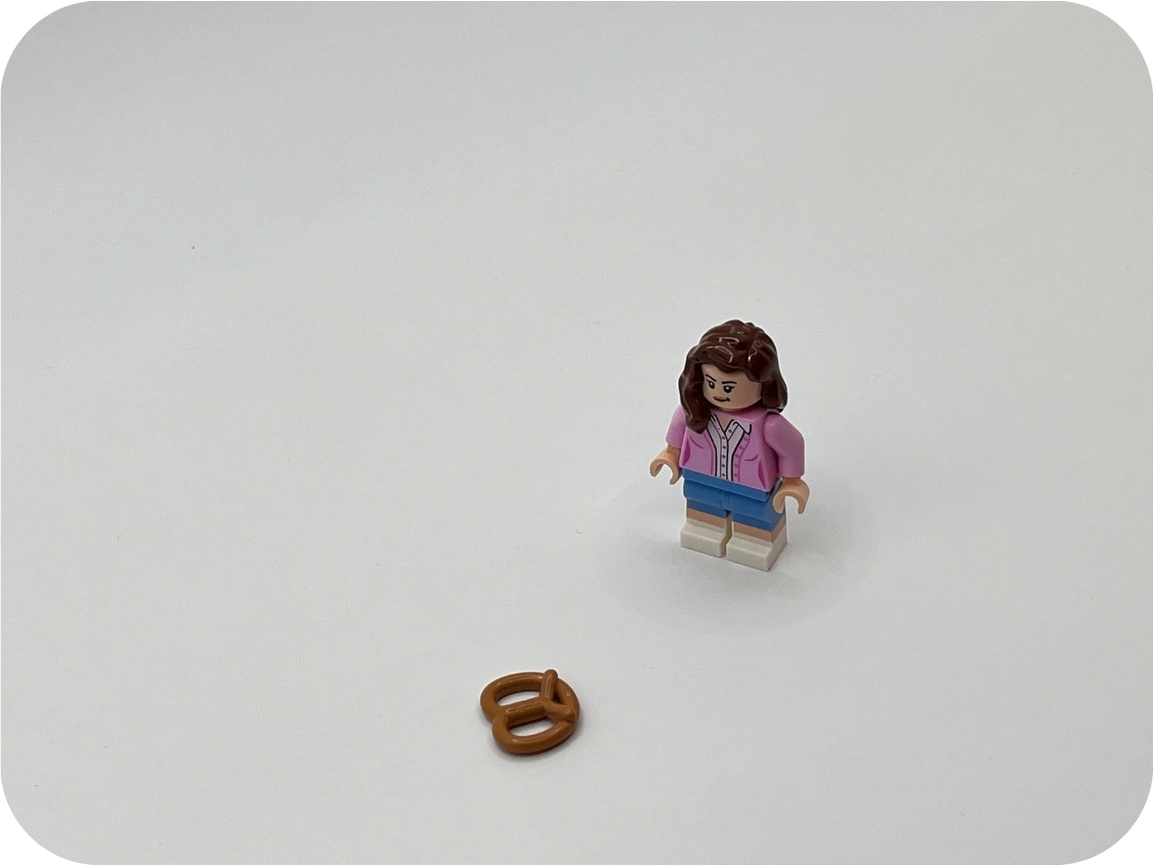}
\caption{\centering Partition 1}
\label{fig:lego-level-1}
\vspace{10pt}
\end{subfigure}
\begin{subfigure}{0.3\linewidth}
\includegraphics[width=\linewidth]{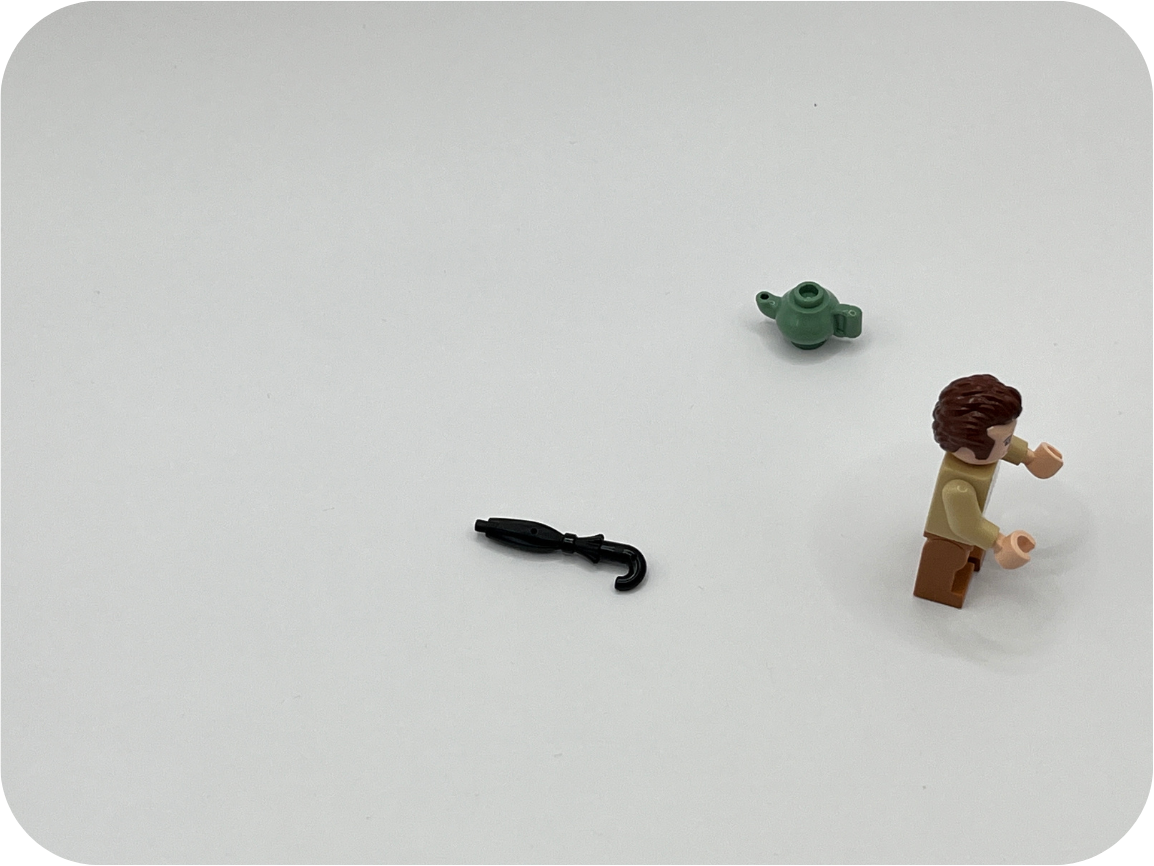}
\caption{\centering Partition 2}
\label{fig:lego-level-2}
\vspace{10pt}
\end{subfigure}
\begin{subfigure}{0.3\linewidth}
\includegraphics[width=\linewidth]{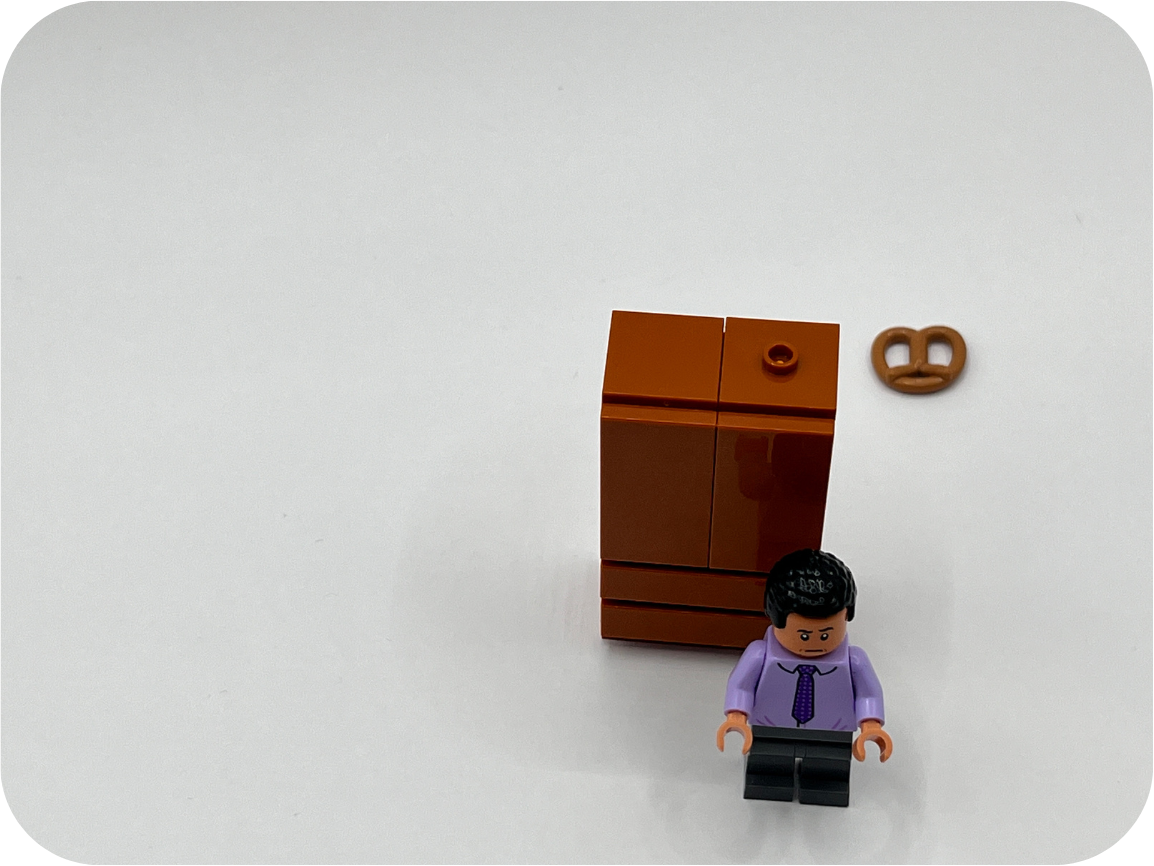}
\caption{\centering Partition 3}
\label{fig:lego-level-3}
\end{subfigure}
\begin{subfigure}{0.3\linewidth}
\includegraphics[width=\linewidth]{images/dataset_photos/edit_14.png}
\caption{\centering Partition 4}
\label{fig:lego-level-4}
\end{subfigure}
\begin{subfigure}{0.3\linewidth}
\includegraphics[width=\linewidth]{images/dataset_photos/edit_15.png}
\caption{\centering Partition 5}
\label{fig:lego-level-5}
\end{subfigure}
\caption{Examples of photos from each partition of the \legodataset{} dataset.
}
\label{fig:lego-levels}
\end{figure}

The dataset is split into six groups, varying by difficulty:

\begin{itemize}
    \item Photos containing exactly one Lego piece, a Lego figurine, or an object. Those photos do not test perspective taking but the object detection capabilities of a model.
    \item Photos that contain exactly two Lego pieces, one figurine, and one object. This group is the easiest kind of a perspective taking task.
    \item Photos containing exactly three Lego pieces, one figurine, and two objects. This group is slightly more challenging than the first one since the model has to actually prove that it is capable of telling objects from one another.
    \item Photos containing exactly three Lego pieces, one figurine, one object, and a wardrobe that is or is not in the line of sight of the figurine. This is more challenging than the previous one because it requires that the model check if something would be in the line of sight of a person.
    \item Photos containing exactly four Lego pieces, two figurines, one object, and a wardrobe that is or is not in the line of sight of the figurine. This is more challenging than the previous one because it requires that the model identifies and takes the perspective of a specific person.
    \item Photos containing exactly four Lego pieces, one figurine, two objects, and a wardrobe that is or is not in the line of sight of the figurine. This is more challenging than the previous one because it requires that the model identifies the correct object.
\end{itemize}
We provide an example from each partition shown in Figure \ref{fig:lego-levels}. Additionally, the prompts used for the perspective taking and object detection tasks are detailed in Table \ref{tab:isle_brick_prompts}.

\begin{table}[ht]
\centering
\caption{Questions used for Isle-Brick dataset}
\small
\begin{tabular}{>{\raggedright\arraybackslash}p{0.45\linewidth} >{\raggedright\arraybackslash}p{0.45\linewidth}}
\toprule
\textbf{Perspective Taking Task} & \textbf{Object Detection} \\
\midrule
\textbf{\texttt{Assuming the Lego minifigure with a \{attribute\} has the ability to see, can it see the \{object\} in the picture? Respond solely with Yes or No.}} & \textbf{\texttt{Is the \{object\} present in the image? Respond solely with Yes or No.}} \\
\texttt{Options:} & \texttt{Options:} \\
\texttt{A. Yes} & \texttt{A. Yes} \\
\texttt{B. No} & \texttt{B. No} \\
\bottomrule
\end{tabular}
\label{tab:isle_brick_prompts}
\end{table}

\subsection{\dotsdataset{} Dataset}
The \dotsdataset{} dataset contains \textbf{100} pairs of images and questions for testing the perspective taking capabilities. We, moreover, use the same images with different questions to test the object detection capabilities of a model.

\begin{figure}[h]
\centering
\begin{subfigure}{0.3\linewidth}
\includegraphics[width=\linewidth]{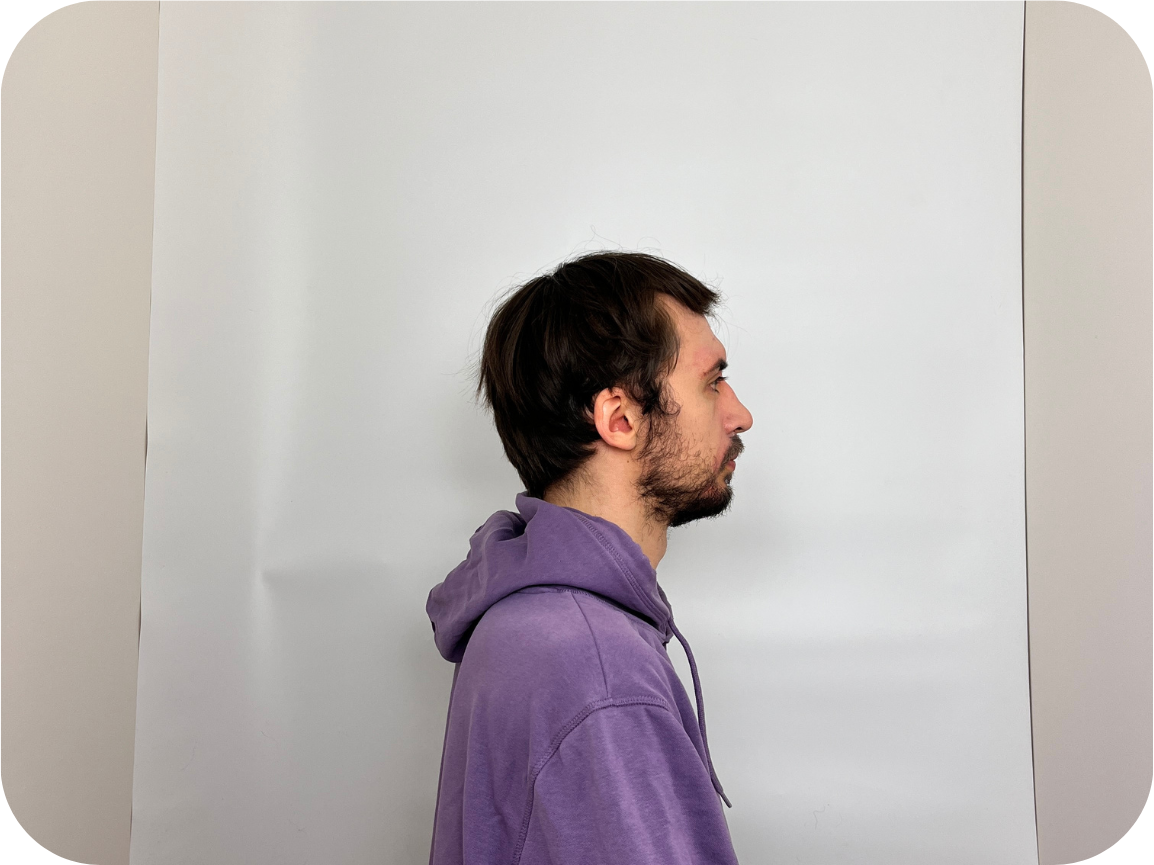}
\caption{\centering Partition 0}
\label{fig:f2d-level-0}
\vspace{10pt}
\end{subfigure}
\begin{subfigure}{0.3\linewidth}
\includegraphics[width=\linewidth]{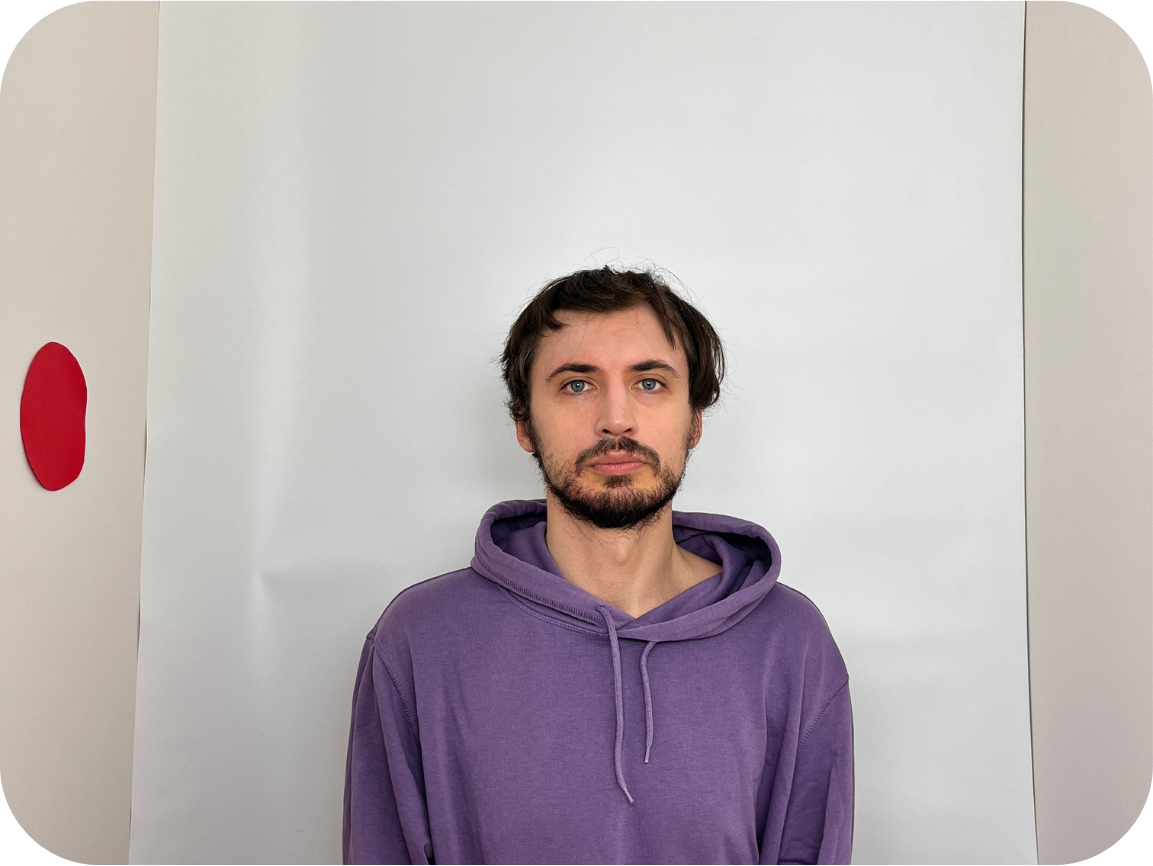}
\caption{\centering Partition 1}
\label{fig:f2d-level-1}
\vspace{10pt}
\end{subfigure}
\begin{subfigure}{0.3\linewidth}
\includegraphics[width=\linewidth]{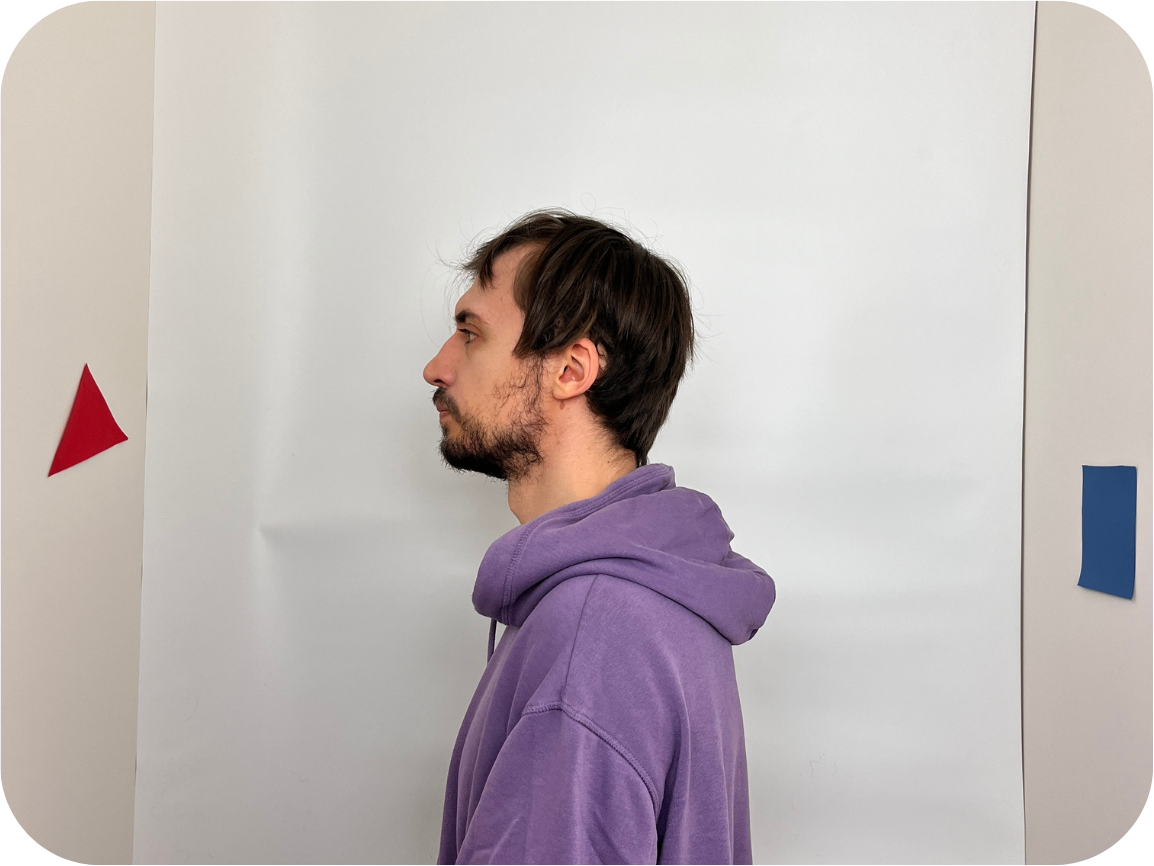}
\caption{\centering Partition 2}
\label{fig:f2d-level-2}
\vspace{10pt}
\end{subfigure}
\begin{subfigure}{0.3\linewidth}
\includegraphics[width=\linewidth]{images/dataset_photos/edit_7.png}
\caption{\centering Partition 3}
\label{fig:f2d-level-3}
\end{subfigure}
\begin{subfigure}{0.3\linewidth}
\includegraphics[width=\linewidth]{images/dataset_photos/edit_8.png}
\caption{\centering Partition 4}
\label{fig:f2d-level-4}
\end{subfigure}
\caption{Examples of photos from each partition of the \dotsdataset{} dataset.
}
\label{fig:f2d-levels}
\end{figure}
\label{app:f2d}

The \dotsdataset{} dataset is divided into the following five partitions depending on the difficulty:
\begin{itemize}
    \item Photos that contain either figures on walls or a person that is turned in some direction
    \item Photos that contain a single figure on a wall and a person that is turned in some direction
    \item Photos that contain two figures on walls and a person that is turned in some direction
    \item Photos that contain three figures on walls on a wall and a person that is turned in some direction
    \item Photos that contain three figures on walls on a wall and a person that is turned in some direction
\end{itemize}
We provide an example from each partition shown in Figure \ref{fig:f2d-levels}. Additionally, the prompts used for the perspective taking and object detection tasks are detailed in Table \ref{tab:isle_f2d_prompts}.

\begin{table}[ht]
\centering
\small
\caption{Questions used for Isle-Dots dataset}
\begin{tabular}{>{\raggedright\arraybackslash}p{0.45\linewidth} >{\raggedright\arraybackslash}p{0.45\linewidth}}
\toprule
\textbf{Perspective Taking Task} & \textbf{Object Detection} \\
\midrule
\textbf{\texttt{In the picture, how many \{color\}\{object\} does the man with the purple hoodie see? Consider only the dots that are directly visible within the picture's frame. Respond solely with A or B.}} & \textbf{\texttt{In the picture, how many \{color\}\{object\} are there? Consider only the rectangles that are directly visible within the picture's frame. Respond solely with A or B.}} \\
\texttt{Options:} & \texttt{Options:} \\
\texttt{A. \{Option 1\}} & \texttt{A. \{Option 1\}} \\
\texttt{B. \{Option 2\}} & \texttt{B. \{Option 2\}} \\
\bottomrule
\end{tabular}
\label{tab:isle_f2d_prompts}
\end{table}

\subsection{Data Labelling}
Our datasets were labelled by the authors, and only the data points where all authors agreed were included in the datasets. This ensures that our datasets do not contain controversial data points for adult humans.
Hence, we would expect a perfect vision language model to respond to them correctly.



\newpage

\end{document}